\documentclass{article}
\usepackage{spconf,amsmath,graphicx,hyperref}
\usepackage{multirow}
\usepackage{CJKutf8}
\usepackage{listings}
\usepackage{graphicx}
\usepackage{cite}
\usepackage{amsmath,amssymb,amsfonts}
\usepackage{algorithmic}
\usepackage{graphicx}
\usepackage{textcomp}
\usepackage{xcolor}
\usepackage{enumitem}
\definecolor{myblue}{RGB}{169, 196, 235}
\definecolor{myyellow}{RGB}{255, 205, 153}
\definecolor{mygreen}{RGB}{185, 225, 165}
\def\BibTeX{{\rm B\kern-.05em{\sc i\kern-.025em b}\kern-.08em
    T\kern-.1667em\lower.7ex\hbox{E}\kern-.125emX}}
    
\title{CHAIN OF CORRECTION FOR FULL-TEXT SPEECH RECOGNITION WITH LARGE LANGUAGE MODELS}
    
\name{Zhiyuan Tang${^\dagger}$, Dong Wang${^\ddagger}$, Zhikai Zhou${^\dagger}$, Yong Liu${^\dagger}$, Shen Huang${^\dagger}$, Shidong Shang${^\dagger}$}
\address{
    ${^\dagger}$Tencent Ethereal Audio Lab, Tencent, China\\
    ${^\ddagger}$Center for Speech and Language Technologies, BNRist, Tsinghua University, China\\
    atomtang@tencent.com, wangdong99@mails.tsinghua.edu.cn
    }
    
\begin{document}

\maketitle

\begin{abstract}
Full-text error correction with Large Language Models (LLMs) for Automatic Speech Recognition (ASR) is attracting increased attention for its ability to address a wide range of error types, such as punctuation restoration and inverse text normalization, across long context. However, challenges remain regarding stability, controllability, completeness, and fluency. To mitigate these issues, this paper proposes the Chain of Correction (CoC), which uses a multi-turn chat format to correct errors segment by segment, guided by pre-recognized text and full-text context for better semantic understanding. Utilizing the open-sourced ChFT dataset, we fine-tune a pre-trained LLM to evaluate CoC's performance. Experiments show that CoC significantly outperforms baseline and benchmark systems in correcting full-text ASR outputs. We also analyze correction thresholds to balance under-correction and over-rephrasing, extrapolate CoC on extra-long ASR outputs, and explore using other types of information to guide error correction.

\end{abstract}

\begin{keywords}
speech recognition, full text, error correction, large language model
\end{keywords}

\section{Introduction}
Automatic Speech Recognition (ASR) systems are integral to various applications like voice search, commands, and transcription services. Nonetheless, factors such as background noise, speaker accents, and audio quality often impair ASR effectiveness, leading to errors that undermine downstream applications. Thus, error correction is essential for improving ASR accuracy.

A common approach employs a language model (LM) to rescore N-best hypotheses from ASR, selecting the candidate with the lowest perplexity~\cite{mikolov2010recurrent,arisoy2015bidirectional,shin2019effective,yang2021multi,yu2023low}. However, this method overlooks useful information in other hypotheses. Alternatively, merging N-best hypotheses can generate a more accurate prediction~\cite{guo2019spelling,hu2020deliberation,leng2021fastcorrect,hu2022improving,ma2023n,hu2023scaling}.

Recently, large language models (LLMs) have been used for generative error correction~\cite{chen2023generative,chen2024hyporadise,tang24_interspeech}, directly producing transcriptions from N-best hypotheses. However, most work focuses on single sentences due to ASR training data limitations, restricting document-level context modeling. Moreover, utterance-level correction is computationally intensive, requiring multiple hypotheses per utterance.

Our previous work~\cite{tang2025full} proposed a novel benchmark specifically designed for generative error correction in full-text documents. This benchmark aimed to thoroughly explore and evaluate the potential of LLMs in identifying and correcting a wide range of errors within full text. By full text, we refer to the entire text of a document, such as an article, news report, or conversation transcript. This also covers error types from punctuation restoration and inverse text normalization (ITN), making the task comprehensive. Specifically, the work introduced the Chinese Full-text Error Correction Dataset (ChFT) and designed various prompts for error correction with LLMs, among which the JSON output format of error-correction pairs was found to be the most effective, compact, and controllable. However, this method still has several limitations, including: 
1) incomplete error discovery, as error words were frequently missed;
2) lack of fluency, as correction was applied by replacing the error word with the corrected word, which overlooked the overall fluency of the text; and 
3) error confusion, as the method merely identified the errors but not their exact positions, leading to confusion or over-replacement during the correction process.

\begin{figure*}[ht]
    \centering
    \includegraphics[trim=20 10 30 20,clip,width=\linewidth]{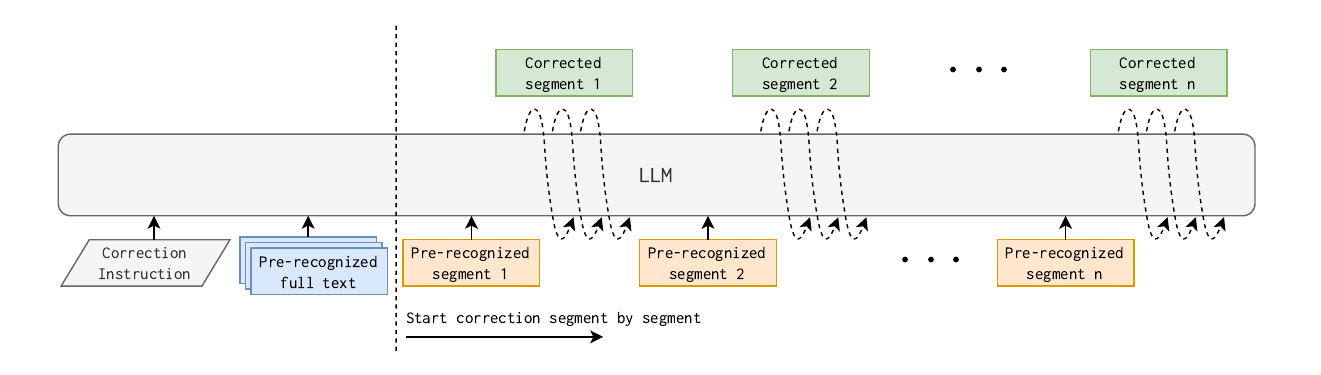}
    \caption{The Chain of Correction (CoC) paradigm for full-text ASR error correction with LLMs. All pre-recognized segments constitute the full text.}
    \label{fig:coc}
\end{figure*}

To address these challenges and other limitations, this paper proposes the Chain of Correction (CoC) for full-text error correction with LLMs, which corrects errors segment by segment using their pre-recognized text as guidance within a regular multi-turn chat format. The CoC model also employs the pre-recognized full text as context, enabling the model to be more aware of global semantics and have an overall sketch of the full text. Using the open-sourced full-text error correction dataset ChFT, we fine-tune a pre-trained LLM to evaluate its performance with the CoC method. We further examine how to establish the correction threshold to balance under-correction and over-rephrasing. Additionally, we assess the CoC model's performance with extra-long ASR outputs and explore the potential of using other types of information to guide the error correction process.

\section{Chain of Correction}
\label{sec:method}
The Chain of Correction (CoC) is a multi-turn chat-based paradigm for full-text error correction with LLMs, as illustrated in Figure~\ref{fig:coc}. Initially, the pre-recognized full text is provided to the LLM as context along with a pre-defined correction instruction. The full text is then segmented, with each segment comprising a few sentences, and the LLM is instructed to correct these segments one by one, each time using the pre-recognized segment as guidance. The segments are corrected in a chain-like manner, with corrected segments from previous turns serving as additional context for subsequent turns. Compared to the previous method~\cite{tang2025full}, which used JSON output format of error-correction pairs, the CoC offers several advantages:
\begin{itemize}[itemsep=0pt, parsep=0pt, topsep=0pt, partopsep=0pt]
    \item Stability: Unlike refreshing the entire text all at once, especially for extra-long text that often leads to hallucination or over-rephrasing, CoC's segment guidance ensures a more stable correction process. The model focuses only on small segments, irrespective of output length limitation during LLM inference.
    \item Controllability: The segment-by-segment approach allows for more flexible control of the correction process, such as frequently checking the degree of over-rephrasing at a segment level, making acceptance or rejection of corrections more manageable and timely. This also keeps the original segment orders intact for better alignment with the audio.
    \item Completeness: CoC rolls over the entire segment using pre-recognized text as guidance, making error exposure easier. It avoids providing the position of errors in the text, whose inaccuracy may lead to confusion or over-correction.
    \item Fluency: Rather than retrieving and replacing exact error words with corrected ones, correcting the segment as a re-generation process word by word naturally ensures greater text fluency, fully leveraging the LLM's next-token prediction capabilities.
\end{itemize}

\begin{figure}[htbp]
    \centering
    \includegraphics[trim=25 25 25 20,clip,width=\linewidth]{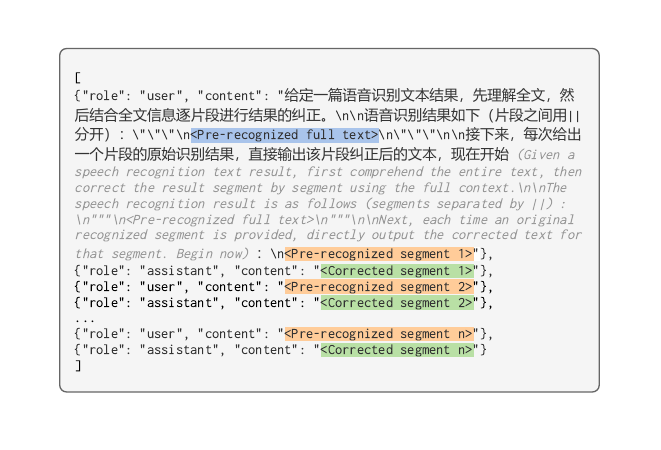}
    \caption{Message template for Chain of Correction.
    The {\color{gray}{\textit{gray}}} part is for translation only.
    The {\colorbox{myblue}{blue}} block represents the pre-recognized full text as context.
    The {\colorbox{myyellow}{yellow}} and {\colorbox{mygreen}{green}} blocks are the pre-recognized segments to be corrected and the corrected ones, respectively.}
    \label{fig:prompt}
\end{figure}

\subsection{Prompt Design}
As shown in Figure~\ref{fig:prompt}, CoC uses a multi-turn chat format. The first `user' turn gives the correction instruction and pre-recognized full text as context. In each turn, the `user' provides a segment to be corrected, and the `assistant' returns the corrected segment.

\subsection{Correction Threshold}
To balance under-correction and over-rephrasing, we introduce a Correction Threshold controlling correction strength and higher values mean stronger correction. After each correction, the Error Rate (see Section~\ref{sec:experiments}) between the original and corrected segment is computed. If it does not exceed the threshold, the correction is accepted; otherwise, it is rejected or further revision is requested.

\section{Experiments}
\label{sec:experiments}
The ChFT dataset~\cite{tang2025full} is used for experiments. It was generated using a text-to-speech (TTS) and ASR pipeline, and contains 41,651 articles with reference-hypothesis pairs for training across diverse domains. Three types of test sets, homogeneous, hard, and up-to-date, each with thousands of articles, evaluate model generalization in different scenarios. Message formats are prepared as shown in Figure~\ref{fig:prompt}. Hypothesis and reference texts are aligned, then the hypothesis is split into segments of 1-5 sentences, ending with terminal punctuation (period, question mark, or exclamation mark). References are split accordingly to form segment-reference pairs.

We fine-tune all parameters of an internal pre-trained LLM (Hunyuan-7B-Dense-Pretrain-256k-V2-241208, a 7B parameter model with 256k context length) on ChFT training messages using the CoC paradigm, employing 16 NVIDIA A100 GPUs for approximately one epoch. Following ASR metrics, the Character/Word Error Rate (ER) is used to assess performance for pure Chinese or code-switched English (CS-English). The ER is defined as:

{\footnotesize
$$
ER = \frac{S+D+I}{N}
$$
}

\noindent where \( N \) is the total number of characters/words in the reference, and \( S \), \( D \), and \( I \) are the numbers of substitutions, deletions, and insertions, respectively. For clearer comparison, we also utilized Error Rate Reduction (ERR):

{\footnotesize
$$
ERR = \frac{ER_{\text{fine-tuned}} - ER_{\text{baseline}}}{ER_{\text{baseline}}}
$$
}

\noindent where \( ER_{\text{fine-tuned}} \) and \( ER_{\text{baseline}} \) are the ERs from the fine-tuned LLM and baseline ASR, respectively. The same evaluation is used for punctuation and ITN correction.

\subsection{Performance on Full-text Error Correction}
\label{sec:performance}

The performance of the fine-tuned LLM on full-text error correction is shown in Table~\ref{tab:res-coc}, together with baseline and benchmarking systems from previous work~\cite{tang2025full}. To further assess the CoC paradigm, we also used DeepSeek-R1~\cite{guo2025deepseek} (671B parameters) to correct the up-to-date test set for comparison.
The correction threshold is set to 0.3 for all experiments, and other settings are discussed below.

\begin{table}[htbp]
    \caption{Results of Chain of Correction (CoC) on ChFT test sets.}
    \label{tab:res-coc}
    \resizebox{0.5\textwidth}{!}{

\begin{tabular}{|l|l|lllll|}
\hline
\multirow{2}{*}{\textbf{Test set}} & \multicolumn{1}{c|}{\multirow{2}{*}{\textbf{System}}} & \multicolumn{5}{c|}{\textbf{ER\%}$\downarrow$ $_{ERR\%\downarrow}$}                                                                                                                     \\ \cline{3-7} 
                                   & \multicolumn{1}{c|}{}                                 & \multicolumn{1}{l|}{Mandarin} & \multicolumn{1}{l|}{Punctuation} & \multicolumn{1}{l|}{ITN}   & \multicolumn{1}{l|}{CS-English} & \multicolumn{1}{l|}{Overall} \\ \hline
\multirow{3}{*}{Homogeneous}       & Baseline                                              & \multicolumn{1}{l|}{6.16}     & \multicolumn{1}{l|}{67.18}       & \multicolumn{1}{l|}{45.17} & \multicolumn{1}{l|}{80.53}   & 12.61                        \\ \cline{2-7} 
                                   & seg\_json\cite{tang2025full}                                             & \multicolumn{1}{l|}{4.78$_{\color{teal} -22.40}$}     & \multicolumn{1}{l|}{37.68$_{\color{teal} -43.91}$}       & \multicolumn{1}{l|}{29.23$_{\color{teal} -35.29}$} & \multicolumn{1}{l|}{56.41$_{\color{teal} -29.95}$}   & 8.41$_{\color{teal} -33.31}$                         \\ \cline{2-7} 
                                   & CoC                                                   & \multicolumn{1}{l|}{\textbf{4.06$_{\color{teal} -34.09}$}}     & \multicolumn{1}{l|}{\textbf{32.32$_{\color{teal} -51.89}$}}       & \multicolumn{1}{l|}{\textbf{18.63$_{\color{teal} -58.76}$}} & \multicolumn{1}{l|}{\textbf{46.97$_{\color{teal} -41.67}$}}   & \textbf{7.03$_{\color{teal} -44.25}$}                         \\ \hline
\multirow{3}{*}{Hard}              & Baseline                                              & \multicolumn{1}{l|}{19.77}    & \multicolumn{1}{l|}{70.55}       & \multicolumn{1}{l|}{54.89} & \multicolumn{1}{l|}{89.08}   & 25.24                        \\ \cline{2-7} 
                                   & seg\_json\cite{tang2025full}                                             & \multicolumn{1}{l|}{18.85$_{\color{teal} -4.65}$}    & \multicolumn{1}{l|}{49.80$_{\color{teal} -29.41}$}       & \multicolumn{1}{l|}{45.00$_{\color{teal} -18.02}$} & \multicolumn{1}{l|}{71.99$_{\color{teal} -19.19}$}   & 22.36$_{\color{teal} -11.41}$                        \\ \cline{2-7} 
                                   & CoC                                                   & \multicolumn{1}{l|}{\textbf{17.80$_{\color{teal} -9.96}$}}    & \multicolumn{1}{l|}{\textbf{46.12$_{\color{teal} -34.63}$}}       & \multicolumn{1}{l|}{\textbf{38.79$_{\color{teal} -29.33}$}} & \multicolumn{1}{l|}{\textbf{69.39$_{\color{teal} -22.10}$}}   & \textbf{20.97$_{\color{teal} -16.92}$}                        \\ \hline
\multirow{3}{*}{Up-to-date}        & Baseline                                              & \multicolumn{1}{l|}{5.97}     & \multicolumn{1}{l|}{55.68}       & \multicolumn{1}{l|}{38.23} & \multicolumn{1}{l|}{89.85}   & 10.80                        \\ \cline{2-7} 
& DeepSeek-R1                                           & \multicolumn{1}{l|}{6.92$_{\color{teal}  15.91}$}           & \multicolumn{1}{l|}{49.40$_{\color{teal} -11.28}$}          & \multicolumn{1}{l|}{37.09$_{\color{teal} -2.98}$}          & \multicolumn{1}{l|}{77.21$_{\color{teal} -14.07}$}          & 11.09$_{\color{teal} 2.69}$          \\ \cline{2-7} 
                                   & seg\_json\cite{tang2025full}                                             & \multicolumn{1}{l|}{5.13$_{\color{teal} -14.07}$}     & \multicolumn{1}{l|}{32.78$_{\color{teal} -41.13}$}       & \multicolumn{1}{l|}{19.14$_{\color{teal} -49.93}$} & \multicolumn{1}{l|}{52.38$_{\color{teal} -41.70}$}   & 7.75$_{\color{teal} -28.24}$                         \\ \cline{2-7} 
                                   & CoC                                                   & \multicolumn{1}{l|}{\textbf{4.19$_{\color{teal} -29.82}$}}     & \multicolumn{1}{l|}{\textbf{28.65$_{\color{teal} -48.55}$}}       & \multicolumn{1}{l|}{\textbf{16.71$_{\color{teal} -56.29}$}} & \multicolumn{1}{l|}{\textbf{46.68$_{\color{teal} -48.05}$}}   & \textbf{6.51$_{\color{teal} -39.72}$}                         \\ \hline
\end{tabular}

    }
\end{table}

The results in Table~\ref{tab:res-coc} show that the CoC paradigm significantly outperforms the baseline and benchmarking systems on all test sets and error types. Notably, on the up-to-date test set (articles published after July 1, 2024, and excluded from pre-training), CoC still achieves substantial error reduction. On the hard test set with challenging conditions like babble noise, CoC achieves about 9.96\% improvement in Mandarin error correction. DeepSeek-R1's performance is less impressive, likely due to overthinking the task and generating overly rephrased or verbose outputs that ignore the prescribed format. This highlights the value of task-specific fine-tuning with smaller LLMs.

\subsection{Correction Threshold Setting}
\label{sec:correct_threshold}
\begin{figure}[htbp]
    \centering
    \includegraphics[trim=0 10 0 0,clip,width=0.8\linewidth]{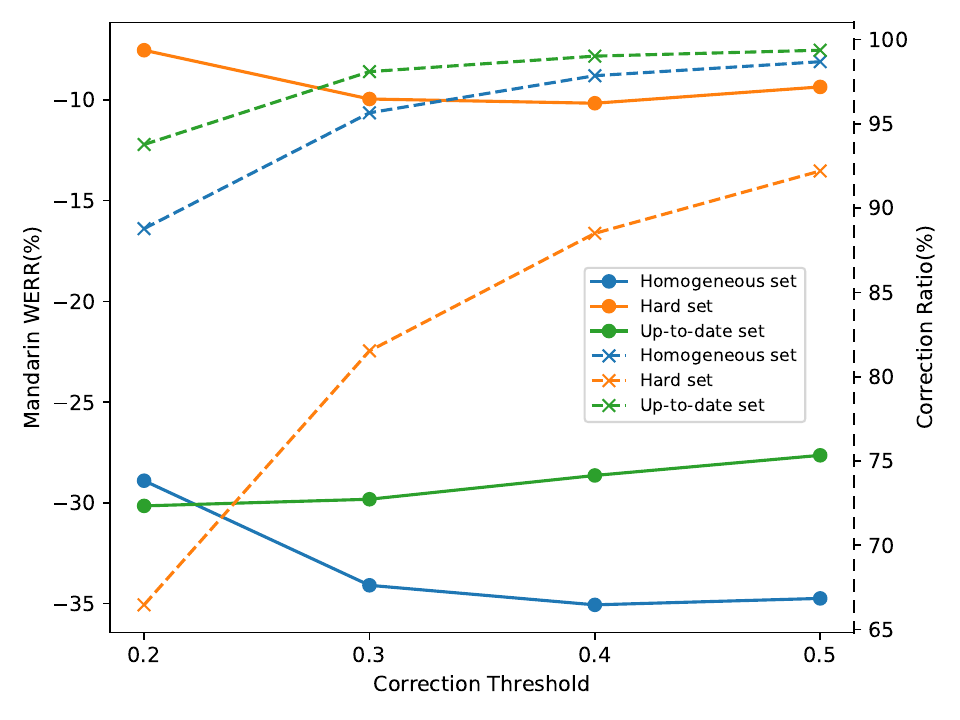}
    \caption{The trend of Mandarin ERR and correction ratio with different Correction Threshold values. The solid line represents the ERR, and the dashed line indicates the correction ratio.}
    \label{fig:threshold}
\end{figure}

The Correction Threshold controls the model's correction strength: a larger threshold increases correction strength but may cause over-rephrasing, while a smaller threshold may lead to under-correction. We evaluate CoC's performance with different Correction Thresholds on the three test sets. Figure~\ref{fig:threshold} shows the Mandarin ERR as the threshold increases from 0.2 to 0.5, along with the correction ratio (the proportion of accepted corrections out of all segments). If a correction is rejected, the original segment is retained and correction moves to the next segment.

Raising the Correction Threshold increases the correction ratio and generally improves performance. However, too high a threshold can cause over-rephrasing and slightly reduce performance. A threshold of 0.3 or 0.4 provides an optimal balance on the ChFT dataset, so 0.3 is used in other experiments.

\subsection{Extrapolation on Extra-long Context}
\label{sec:extremely_long}
The context of the ChFT dataset has a maximum length of around 6k characters, corresponding to approximately 4k tokens after tokenization, whereas the pre-trained base model used in this work supports a context length of up to 256k tokens, far exceeding the length of the ChFT dataset.

\begin{table}[htbp]
    \centering
    \caption{Statistics of the extra-long test set.}
    \label{tab:extremely_long_stat}
    \resizebox{0.3\textwidth}{!}{

\begin{tabular}{|l|l|l|l|}
\hline
                 & Minimum & Maximum & Average \\ \hline
Character length & 12,000        & 80,697        &23,984         \\ \hline
Audio duration (min)  & 11.88         & 248.50         & 72.99         \\ \hline
\end{tabular}
    }
\end{table}

To evaluate the CoC model with extra-long context, we randomly selected 100 long articles from the IndustryCorpus2 dataset\footnote{https://huggingface.co/datasets/BAAI/IndustryCorpus2}, each with at least 12k characters (about 8k tokens after tokenization). Following the same TTS-ASR pipeline as in~\cite{tang2025full}, we created an extra-long test set totaling 59,147 segments, divided via Voice Activity Detection (VAD) in the ASR system. This test set is also included in the ChFT dataset\footnote{https://huggingface.co/datasets/tzyll/ChFT}. Table~\ref{tab:extremely_long_stat} shows statistics of character length and audio duration, and the maximum token length is about 54k, matching audio duration of up to 4 hours. All messages for a single article sum to roughly 3 times the article length, because the model corrects the article (x1), uses pre-recognized segments as guidance (x1), and uses the pre-recognized full text as context (x1). Thus, the maximum token count per article's messages in this test set can approach 160k.

\begin{table}[h]
    \centering
    \caption{Results of Chain of Correction (CoC) on extra-long context.}
    \label{tab:extremely_long}
    \resizebox{0.4\textwidth}{!}{

\begin{tabular}{|l|lllll|}
\hline
\multicolumn{1}{|c|}{\multirow{2}{*}{\textbf{System}}} & \multicolumn{5}{c|}{\textbf{ER\%}$\downarrow$ $_{ERR\%\downarrow}$}                                                                                                                     \\ \cline{2-6} 
\multicolumn{1}{|c|}{}                                 & \multicolumn{1}{l|}{Mandarin} & \multicolumn{1}{l|}{Punctuation} & \multicolumn{1}{l|}{ITN}   & \multicolumn{1}{l|}{CS-English} & \multicolumn{1}{l|}{Overall} \\ \hline
Baseline                                               & \multicolumn{1}{l|}{5.14}     & \multicolumn{1}{l|}{61.64}       & \multicolumn{1}{l|}{46.38} & \multicolumn{1}{l|}{76.90}   & 11.78                        \\ \hline
CoC                                                    & \multicolumn{1}{l|}{4.19$_{\color{teal} -18.48}$}     & \multicolumn{1}{l|}{56.42$_{\color{teal} -8.47}$}       & \multicolumn{1}{l|}{32.44$_{\color{teal} -30.06}$} & \multicolumn{1}{l|}{62.29$_{\color{teal} -19.00}$}   & 10.18$_{\color{teal} -13.58}$                        \\ \hline
\end{tabular}

    }
\end{table}

The results in Table~\ref{tab:extremely_long} illustrate that the CoC model still achieves significant error reduction on the extra-long test set. Specifically, the ER is reduced by 18.48\% compared to the baseline model for Mandarin text, with consistent reductions in other types of errors, demonstrating the CoC model's potential with extra-long context.

\subsection{Pinyin as Guidance}
\label{sec:pinyin}
Typically, the pre-recognized segment guides the model in error correction. Here, we explore using alternative representations, such as pinyin, to guide this process. As a phonetic representation of Chinese, pinyin may offer the model different correction cues. Previous work shows pinyin can enhance ASR correction~\cite{tang24_interspeech,li2025large}. We use pinyin representations of pre-recognized segments as guidance in the CoC model.

Table~\ref{tab:pinyin} shows that pinyin guidance improves correction performance, though less than original hypothesis guidance. This indicates the potential for other representation forms, such as discrete speech tokens from WavLM~\cite{chen2022wavlm}, which can facilitate the integration of speech recognition and error correction.

\begin{table}[]
    \centering
    \caption{Results of Chain of Correction (CoC) with pinyin as guidance on the ChFT dataset.}
    \label{tab:pinyin}
    \resizebox{0.5\textwidth}{!}{

\begin{tabular}{|l|l|lllll|}
\hline
\multirow{2}{*}{\textbf{Test set}} & \multicolumn{1}{c|}{\multirow{2}{*}{\textbf{System}}} & \multicolumn{5}{c|}{\textbf{ER\%}$\downarrow$ $_{ERR\%\downarrow}$}                                                                                                                     \\ \cline{3-7} 
                                   & \multicolumn{1}{c|}{}                                 & \multicolumn{1}{l|}{Mandarin} & \multicolumn{1}{l|}{Punctuation} & \multicolumn{1}{l|}{ITN}   & \multicolumn{1}{l|}{CS-English} & \multicolumn{1}{l|}{Overall} \\ \hline
\multirow{3}{*}{Homogeneous}       & Baseline                                              & \multicolumn{1}{l|}{6.16}     & \multicolumn{1}{l|}{67.18}       & \multicolumn{1}{l|}{45.17} & \multicolumn{1}{l|}{80.53}   & 12.61                        \\ \cline{2-7} 
                                   & CoC (hyp guided)                                      & \multicolumn{1}{l|}{4.06$_{\color{teal} -34.09}$}     & \multicolumn{1}{l|}{32.32$_{\color{teal} -51.89}$}       & \multicolumn{1}{l|}{18.63$_{\color{teal} -58.76}$} & \multicolumn{1}{l|}{46.97$_{\color{teal} -41.67}$}   & 7.03$_{\color{teal} -44.25}$                         \\ \cline{2-7} 
                                   & CoC (pinyin guided)                                   & \multicolumn{1}{l|}{4.24$_{\color{teal} -31.17}$}     & \multicolumn{1}{l|}{33.75$_{\color{teal} -49.76}$}       & \multicolumn{1}{l|}{20.06$_{\color{teal} -55.59}$} & \multicolumn{1}{l|}{49.41$_{\color{teal} -38.64}$}   & 7.35$_{\color{teal} -41.71}$                         \\ \hline
\multirow{3}{*}{Hard}              & Baseline                                              & \multicolumn{1}{l|}{19.77}    & \multicolumn{1}{l|}{70.55}       & \multicolumn{1}{l|}{54.89} & \multicolumn{1}{l|}{89.08}   & 25.24                        \\ \cline{2-7} 
                                   & CoC (hyp guided)                                      & \multicolumn{1}{l|}{17.80$_{\color{teal} -9.96}$}    & \multicolumn{1}{l|}{46.12$_{\color{teal} -34.63}$}       & \multicolumn{1}{l|}{38.79$_{\color{teal} -29.33}$} & \multicolumn{1}{l|}{69.39$_{\color{teal} -22.10}$}   & 20.97$_{\color{teal} -16.92}$                        \\ \cline{2-7} 
                                   & CoC (pinyin guided)                                   & \multicolumn{1}{l|}{17.95$_{\color{teal} -9.21}$}    & \multicolumn{1}{l|}{46.74$_{\color{teal} -33.75}$}       & \multicolumn{1}{l|}{39.27$_{\color{teal} -28.46}$} & \multicolumn{1}{l|}{69.88$_{\color{teal} -21.55}$}   & 21.17$_{\color{teal} -16.13}$                        \\ \hline
\multirow{3}{*}{Up-to-date}        & Baseline                                              & \multicolumn{1}{l|}{5.97}     & \multicolumn{1}{l|}{55.68}       & \multicolumn{1}{l|}{38.23} & \multicolumn{1}{l|}{89.85}   & 10.80                        \\ \cline{2-7} 
                                   & CoC (hyp guided)                                      & \multicolumn{1}{l|}{4.19$_{\color{teal} -29.82}$}     & \multicolumn{1}{l|}{28.65$_{\color{teal} -48.55}$}       & \multicolumn{1}{l|}{16.71$_{\color{teal} -56.29}$} & \multicolumn{1}{l|}{46.68$_{\color{teal} -48.05}$}   & 6.51$_{\color{teal} -39.72}$                         \\ \cline{2-7} 
                                   & CoC (pinyin guided)                                   & \multicolumn{1}{l|}{4.85$_{\color{teal} -18.76}$}     & \multicolumn{1}{l|}{31.20$_{\color{teal} -43.97}$}       & \multicolumn{1}{l|}{29.15$_{\color{teal} -23.75}$} & \multicolumn{1}{l|}{52.62$_{\color{teal} -41.44}$}   & 7.54$_{\color{teal} -30.19}$                         \\ \hline
\end{tabular}

    }
\end{table}

\section{Discussion}
\label{sec:discussion}
We analyze some correction examples of the CoC model on the ChFT test set. Besides the expected correction of pure text, commonly used punctuation restoration, and ITN, we also find several typical examples that are challenging to correct using common sentence-level methods:

\begin{itemize}[itemsep=0pt, parsep=0pt, topsep=0pt, partopsep=0pt]
    \item VAD Revision: VAD in ASR may lead to early splitting of a segment, resulting in incorrect terminal punctuation. CoC can correct that by removing the erroneous terminal punctuation or replacing it with right non-terminal one.
    \item Special Punctuation Restoration: Besides the common punctuations such as periods and commas, CoC can also correct special punctuation marks like book title markers \begin{CJK*}{UTF8}{gbsn}《》\end{CJK*}, which are specific to Chinese.
    \item Filler Word and Repetition Removal: To enhance the fluency of the text, the CoC model tends to remove simple filler words such as ``\begin{CJK*}{UTF8}{gbsn}呃\end{CJK*}(uh)" and repetitions.
    \item Truecasing: For code-switched English, the CoC model can easily restore the correct casing.
    \item Coreference Resolution: CoC can resolve coreferences in text, such as pronouns and their antecedents. For instance, it can correct ``\begin{CJK*}{UTF8}{gbsn}他\end{CJK*} (he)" to ``\begin{CJK*}{UTF8}{gbsn}她\end{CJK*} (her)," even though they share the same pronunciation in Chinese.
    \item Named Entity Correction: With the context of the full text, the CoC model can correct named entities in the text, such as unusual names of persons or companies.
\end{itemize}

\section{Conclusion}
\label{sec:conclusion}
In this paper, we propose the Chain of Correction (CoC) paradigm for full-text error correction using LLMs. CoC corrects errors segment by segment with pre-recognized text as guidance in a multi-turn chat format. It significantly reduces errors on the ChFT dataset, outperforming baseline and benchmark systems. We also explore Correction Threshold settings, extrapolation with long context, and pinyin as guidance, showing CoC's versatility. Future work will expand CoC to cover more objectives like spoken-to-written conversion and involve diverse context sources such as web search engine information and user history to enhance correction performance.

\bibliographystyle{IEEEbib}
\bibliography{mybib}

\end{document}